\documentclass[journal,twoside,print]{ieeecolor}
\usepackage{tmi}
\usepackage{cite}
\usepackage{amsmath,amssymb,amsfonts}
\usepackage{graphicx}
\usepackage{epstopdf}
\usepackage{textcomp}
\usepackage{algorithm}
\usepackage[noend]{algpseudocode}
\algrenewcommand\algorithmiccomment[1]{\hfill$\triangleright$ {\footnotesize #1}}
\usepackage{booktabs,array,tabularx}
\usepackage{url}
\usepackage[hidelinks]{hyperref}
\newcolumntype{Y}{>{\raggedright\arraybackslash}X}
\makeatletter
\def\ps@headings{\def\@oddhead{}\def\@evenhead{}%
  \def\@oddfoot{\hfil\footnotesize\thepage\hfil}\def\@evenfoot{\hfil\footnotesize\thepage\hfil}}
\def\ps@titlepagestyle{\ps@headings}
\makeatother
\begin{document}
\title{Representation Bias, Correction Transfer, and Resolution Sensitivity in Three-Dimensional Mitochondrial Morphometry}
\author{Farouk Ganiyu Adewumi and Timothy Oladunni
\thanks{Manuscript submitted September 29, 2026. This work was supported by the Marine Biological Laboratory.}
\thanks{F. G. Adewumi and T. Oladunni are with the Department of Computer Science, Morgan State University, Baltimore, MD 21251 USA (e-mail: fagan1@morgan.edu; timothy.oladunni@morgan.edu). Corresponding author: F. G. Adewumi (fagan1@morgan.edu).}}

\maketitle

\begin{abstract}
Quantitative imaging pipelines can produce precise but systematically different measurements of the same object. We present an empirical reliability assessment of three-dimensional mitochondrial morphometry that connects representation bias, a controlled processing intervention, correction transfer, and resolution sensitivity. Using 2,720 development objects from the 3D Mitochondria Shape Library for Optical Microscopy, we find that occupancy-derived volumes exceed reference mesh volumes by 3.665\% on average despite an intraclass correlation coefficient of 0.994. Boundary analysis identifies an outward label displacement of 0.00304 normalized units. In a controlled label-pipeline reimplementation, removing the depth offset reduces volume error in all 55 analyzed objects by a mean of 1.57 percentage points, approximately 45\% of mean reproduced inflation; the source of the remainder is not isolated. A frozen regression using occupancy-derived features reduces median absolute percentage error from 3.481\% to 0.664\% in 2,728 previously unused objects from the same resource. However, its calibrated error bound covers only 92.1\% overall and 49.2\% in a low-occupancy subgroup, demonstrating that accuracy and uncertainty transfer must be evaluated separately. In 550 rat-cortex objects from the MitoEM resource, coarsening in-plane spacing from 8 to 24 nanometers changes median surface area by minus 10.60\% and sphericity by plus 11.76\%, despite a rank correlation of 0.994. These results provide quantitative checks for distinguishing processing-induced descriptor changes from candidate biological differences, without establishing biological invariance or cross-source correction transfer.
\end{abstract}


\section{Introduction}

\IEEEPARstart{M}{itochondrial} morphology is a regulated phenotype shaped by fission, fusion, and interactions with the cellular environment \cite{ref1}. Its relationship with bioenergetics is bidirectional: structural perturbations can accompany changes in respiration, and metabolic interventions can reshape the mitochondrial network \cite{ref2}. A morphological difference therefore has to be interpreted at two levels. An apparent change in volume, surface area, or sphericity can reflect altered organelle structure, but it can also arise from segmentation, meshing, sampling, or resolution.

Studies that pair structure with function make the biological question more tractable: electron microscopy (EM) morphology with cytochrome c oxidase histochemistry in the same tissue \cite{ref3}, live-cell network imaging with biophysical modeling in pancreatic beta cells \cite{ref4}, single-organelle morphology with fluorescent functional probes \cite{ref5}, and shape phenotyping with metabolic measurements \cite{ref6}. Functional readouts carry assay-dependent uncertainty \cite{ref7}; the structural side of such studies needs the same scrutiny.

For computational imaging, the practical problem is deciding whether a descriptor remains comparable after a change in representation or processing: a high correlation does not establish object-level agreement, and a low average prediction error does not establish that an uncertainty bound remains reliable. We address this problem through paired measurements of the same annotated objects, connecting a representation audit to a controlled intervention in a reimplemented labeling pipeline and then evaluating a frozen correction and its error bound separately. A complementary experiment tests descriptor sensitivity to regridding in EM annotations. The reference meshes are computational comparators, not independent measurements of biological truth.

\subsection{Contributions and Scope}
The contribution is an empirical characterization of linked measurement failure modes, with reproducible evidence and practical checks for quantitative imaging pipelines. The four contributions are:
\begin{enumerate}
\item \textbf{Systematic representation bias.} In 2,720 development objects, occupancy-derived volume has a mean bias of $+3.665$\% despite an intraclass correlation coefficient of 0.994; boundary and repeatability analyses separate a displaced label surface from sampling variability.
\item \textbf{A controlled test of a label-processing parameter.} In 55 non-pilot objects, removing the depth offset in a pipeline reimplementation reduces volume error in every object, by 1.57 percentage points on average, leaving the residual bias unresolved.
\item \textbf{Separate evaluation of correction and uncertainty transfer.} A frozen, mesh-free correction reduces median absolute percentage error from 3.481\% to 0.664\% in 2,728 new objects, but its error-bound coverage is only 92.1\% overall and 49.2\% in a low-occupancy subgroup.
\item \textbf{Absolute descriptor sensitivity despite stable ranks.} Regridding 550 rat-cortex objects to 24 nm in-plane spacing changes median surface area by $-10.60$\% and sphericity by $+11.76$\% while preserving sphericity ranks ($\rho=0.994$).
\end{enumerate}
The novelty lies in the linked experimental evidence, not a new regression algorithm or agreement statistic. Correction transfer is evaluated within one resource, and no linked functional measurements are available to test biological invariance.

\section{Related Work}

\subsection{Automated phenotyping tools}

MiNA measures mitochondrial network skeletons \cite{ref31}. MitoGraph reconstructs three-dimensional (3D) surfaces and estimates volume \cite{ref32}. Nellie extends automated organelle analysis to segmentation, tracking, and hierarchical features in 2D and 3D live-cell microscopy \cite{ref33}.

MitoClass assigns network morphologies to fragmented, intermediate, or elongated categories with a convolutional neural network \cite{ref9}. MitoTex provides texture-based characterization \cite{ref10}. These tools produce descriptors from a chosen representation; the agreement of such descriptors across representations of the same object is the question studied here.

\subsection{Shape representations and public resources}

The 3D Mitochondria Shape Library for Optical Microscopy (3DMSL) provides meshes, point clouds, implicit occupancy data, and simulated optical microscopy images for more than 27,000 mitochondrial shapes \cite{ref11}. The upstream focused ion beam scanning electron microscopy (FIB-SEM) resource is archived as EMPIAR-10791 \cite{ref13}. Its originating study investigated liver subcellular architecture and metabolic homeostasis \cite{ref14}. The 3DMSL repository identifies obese (ob/ob) mouse liver as the source \cite{ref20}; its occupancy labels follow the Occupancy Networks preprocessing \cite{ref21}, in which query points are labeled against a watertight surface produced by depth-map fusion \cite{ref22}. MitoEM provides large-scale instance-labeled EM volumes \cite{ref15}, and the subsequent challenge analysis documents the importance of annotation quality and evaluation design \cite{ref16}. Derived representations of one annotation allow controlled processing studies in which object identity is fixed.

\subsection{The measurement gap}

Table~\ref{tab:I} places this study relative to representative approaches; it compares scope and evidence, not benchmark performance. The structure--function studies motivate paired measurements but do not establish a universal mapping from shape to function. The question here is whether a descriptor remains comparable after representation conversion, and whether a correction and its uncertainty bound transfer together.

\begin{table*}[!t]
\caption{Representative approaches and the measurement question addressed by this study.}
\label{tab:I}
\centering
\footnotesize
\setlength{\tabcolsep}{4pt}
\renewcommand{\arraystretch}{1.15}
\begin{tabularx}{\textwidth}{@{}p{.21\textwidth} Y Y@{}}
\toprule
\textbf{Study or resource} & \textbf{Approach and endpoint} & \textbf{Relationship to this study} \\
\midrule
Faitg et al. \cite{ref3}, 2020 & EM morphology with cytochrome c oxidase labeling; structure--function assessment & Motivates paired biological measurements \\
MitoEM \cite{ref15}, 2020; \cite{ref16}, 2023 & Large-scale 3D EM instance segmentation and its evaluation & Supplies real-EM labels; we measure descriptor sensitivity \\
Tseng et al. \cite{ref4}, 2024 & Live-cell network imaging and biophysical modeling; metabolic coupling & Biological coupling; we characterize measurement prerequisites \\
Nellie \cite{ref33}, 2025 & Automated 2D/3D organelle segmentation; morphology and motility features & Produces descriptors whose sensitivity we quantify \\
Ding et al. \cite{ref5}, 2025 & Fluorescent probes with AI organelle analysis; joint functional biomarkers & Multimodal evidence absent from our datasets \\
3DMSL \cite{ref11}, 2026 & Paired 3D shapes and synthetic microscopy resource & Enables comparison of representations of one object \\
Present study & Bias audit, boundary mechanism, frozen correction, subgroup and resolution tests & Quantifies representation sensitivity; no functional inference \\
\bottomrule\end{tabularx}\end{table*}

\section{Measurement Framework}

Let $S_i$ denote the underlying structure of object $i$ and let $T_r$ denote processing into representation $r$. A descriptor is observed as $m_{ir}=g_r(T_r(S_i))$, so $m_{ir}$ and $m_{is}$ may differ even when the object is unchanged. Our experiments hold object identity fixed and vary representation or processing. The raw 3DMSL mesh serves as the computational reference; it is not treated as an error-free biological measurement.

For mesh volume $V_m$ and occupancy volume $V_o$, the signed percentage error is
\begin{equation}
e=100\left(\frac{V_o}{V_m}-1\right).
\end{equation}
We report mean signed error, median absolute percentage error (MdAPE), and root mean square percentage error (RMSE). Spearman rank correlation assesses ordering. Agreement is summarized with the two-way, absolute-agreement, single-measure intraclass correlation coefficient, ICC(A,1) \cite{ref25}, interpreted following \cite{ref18}. ICC depends on between-object variability relative to disagreement; a wide size range can yield a high ICC despite practically important object-level errors. We therefore also report Bland--Altman bias and limits of agreement \cite{ref17} and the percentage error, $100\times1.96\,\mathrm{SD}_{\Delta}/\bar{M}$, where $\bar{M}=n^{-1}\sum_i(V_{mi}+V_{oi})/2$ \cite{ref26}.

For an approximately uniform outward boundary displacement $t$ that is small relative to local geometric dimensions, the first-order volume change and the implied normalized displacement are
\begin{equation}
\Delta V\approx A_m t,\qquad t_{\mathrm{norm}}\approx\frac{V_o-V_m}{A_m s},
\end{equation}
where $A_m$ is mesh area and $s$ is the stored normalization scale. For constant normalized displacement, the corresponding prediction is
\begin{equation}
e\approx100\,t_{\mathrm{norm}}\frac{A_m s}{V_m}
=100\,t_{\mathrm{norm}}\frac{A_m}{V_m^{2/3}}\frac{s}{V_m^{1/3}}.
\end{equation}
Thus, dimensionless shape complexity $A_m/V_m^{2/3}$ alone omits a potentially variable scale-to-size factor. We report its association with error descriptively and additionally test $A_ms/V_m$ as a mechanistic predictor.

\section{Datasets and Experimental Design}

\subsection{Dataset provenance and units of analysis}

We analyzed two public resources. From 3DMSL \cite{ref12} we used the development archive 10\_24553\_27272.zip (IDs 24553--27272; 2,720 objects) and the frozen validation archive 1\_1\_2729.zip (IDs 1--2728; 2,728 objects, none excluded, no ID overlap). From MitoEM-R \cite{ref15} we used the rat-cortex validation slices 400--499 at $30\times8\times8$ nm (1,203 instance IDs, 550 eligible complete objects); training slices 0--29 served only as a technical pilot. The two 3DMSL ZIP files are batches within one resource, not independent biological cohorts. Their IDs are disjoint, and no raw-mesh file is duplicated across batches, but source-animal independence is not established. EMPIAR-10791 is the upstream source of 3DMSL, not a third analyzed dataset.

E4 uses both 3DMSL batches and E8 reuses the E5 objects, so the subset sizes in Table~\ref{tab:III} should not be summed. An initial 40-object MitoEM analysis and an initial 400-object imaging analysis were superseded by the 550-object and 300-object analyses and are not counted as additional evidence.

\subsection{Experiment-to-claim mapping}

The experiment labels in Table~\ref{tab:III} organize completed analyses; they do not imply prospective registration. E8 and the supplementary analyses were added after the primary results (Supplementary Material, Section~S4). Boundary and distribution-shift diagnostics are exploratory analyses that explain or qualify the primary results.

\begin{table*}[!t]
\caption{Experiment-to-claim map.}
\label{tab:III}
\centering
\footnotesize
\setlength{\tabcolsep}{4pt}
\renewcommand{\arraystretch}{1.15}
\begin{tabularx}{\textwidth}{p{.025\textwidth} Y Y Y}
\toprule
\textbf{ID} & \textbf{Experiment} & \textbf{Sample and comparison} & \textbf{Supported claim} \\
\midrule
E1 & Integrity and representation agreement & 2,720 development objects; mesh, occupancy, point cloud & Systematic disagreement despite high agreement \\
E2 & Occupancy sampling repeatability & 400 objects; 10 disjoint query folds; convergence & Numerical precision is separate from offset \\
E3 & Internal correction and calibration & 1,630 train, 540 calibration, 550 test objects & Internal performance after exploration \\
E4 & Frozen transfer and shift diagnostics & 2,728 new objects; original model and bounds & Within-dataset transfer; coverage failures \\
E5 & Label boundary characterization & 60 objects; distance-label logistic fits & Outward label-boundary offset \\
E6 & Native MitoEM representation comparison & 550 objects from one rat-cortex slab & Mask-to-mesh and sampling sensitivity on EM \\
E7 & In-plane resolution sensitivity & Same 550 objects; native, 16, 24 nm grids & Absolute shifts with rank stability \\
E8 & Controlled label-pipeline reimplementation & 55 E5 objects; depth offsets 0, 0.75, 1.5 voxels & Fidelity; effect of removing the offset \\
\bottomrule\end{tabularx}\end{table*}

\subsection{Structural processing and quality control}

The 3DMSL audit verifies the mesh, point-cloud, and occupancy files for each object. Point-cloud and occupancy coordinates are transformed with $\mathbf{x}_{\mathrm{raw}}=s\,\mathbf{x}_{\mathrm{norm}}+\mathbf{loc}$, and bit-packed occupancy labels are unpacked before processing. Each archive stores 100,000 uniformly distributed query points per object in a normalized box of $\pm0.55$ (padding 0.1), with labels computed against the watertight fused mesh \cite{ref20}. The data article reports 10,000 points \cite{ref11}; the released files and build script contain 100,000. The implemented occupancy volume estimate is the fraction of inside-labeled points multiplied by the product of the observed query-coordinate spans in raw units. This empirical-box convention (not the nominal width $1.1s$) is retained for consistency with the saved correction. Mesh volume is the absolute signed volume; mesh area and axis-aligned extents are recorded separately.

Load failures are hard exclusions. Four conditions are advisory flags: nonwatertight meshes; occupancy fractions below 0.02 (the low-occupancy flag); incomplete rendered views; and a raw, untrimmed point-cloud extent that differs from the mesh extent by more than 10\% along any single axis. The flags were defined before any correction model was fitted. Descriptor comparisons of point-cloud extent use the 1st-to-99th-percentile trimmed extent, which suppresses isolated outliers but also removes genuine extremal points. Both 3DMSL batches contained only watertight meshes.

\subsection{Repeatability estimation}

A simple random sample of 400 numerically ordered development IDs is selected without replacement with NumPy seed 0. The same generator subsequently permutes each object's stored queries into ten disjoint folds. The standard deviation of the ten fold-level volume estimates, divided by $\sqrt{10}$, estimates the standard deviation (SD) of the full-sample estimate under the independent-query model. The pairwise repeatability coefficient is $1.96\sqrt{2}\approx2.77$ times this SD, expressed relative to the estimate. It describes differences between two independent numerical estimates and excludes repeat scanning, segmentation, and biological variability. A convergence analysis evaluates subsets from 1,000 to 100,000 queries.

\subsection{Correction and calibration}

The internal split is stratified by mesh-volume decile (seed 2026, numerically ordered IDs; saved split identifiers retained). Training data alone determine the regression coefficients and feature standardization, calibration data the residual bounds, and test data the internal performance summaries. Because earlier whole-batch exploration may have informed feature selection, the internal test is validation after exploration, not prospective confirmation.

Let $B_q$ be the product of coordinate spans over all query points in raw units, $p$ the fraction labeled inside, and $B_{\mathrm{in}}$ the bounding-box volume of only the inside-labeled points. The volume estimate is $V_o=pB_q$. The five predictors are $V_o/B_{\mathrm{in}}$, $p$, $\log V_o$, $\log B_{\mathrm{in}}$, and the ratio of maximum to minimum coordinate span among the inside-labeled points. They describe bounding-box fill, sampling occupancy, scale, and elongation without requiring the reference mesh. Ordinary least squares, a transparent baseline, predicts the signed percentage error, predictions are clipped to $[-50,100]$\%, and the corrected volume is
\begin{equation}
V_{\mathrm{corrected}}=\frac{V_o}{1+\widehat{e}/100}.
\end{equation}
The global correction substitutes the mean training-set percentage error for $\widehat{e}$, and the uncorrected baseline retains $V_o$. Mesh values are used as development targets and for evaluation but are not predictors at deployment.

For each method, calibration uses the absolute corrected-volume percentage error
\begin{equation}
a_i=\left|100\left(\frac{\widehat V_i}{V_{mi}}-1\right)\right|.
\end{equation}
For the occupancy-only correction, the signed residual is $(e_i-\widehat e_i)/(1+\widehat e_i/100)$, rather than $e_i-\widehat e_i$. The saved bound $q$ is the $\lceil(n_{\mathrm{cal}}+1)0.95\rceil$th ordered score: the 514th of 540 calibration scores here. For $q<100$\%, inversion gives a reference-volume prediction interval
\begin{equation}
\left[\frac{\widehat V}{1+q/100},\frac{\widehat V}{1-q/100}\right].
\end{equation}
The split-conformal construction \cite{ref19} provides marginal coverage under the required exchangeability and fitting conditions, not conditional coverage for each subgroup \cite{ref29}. Batch shift can violate exchangeability \cite{ref28}. We report empirical coverage at the saved bound.

Frozen evaluation preserves the development feature means and scales, coefficients, clipping rule, global offset, and calibrated bounds. Model artifacts were hashed before evaluation and rechecked afterward, and no second-batch outcome was used to update the correction. Subsequent second-batch regressions are exploratory descriptions of error associations, not refits. Algorithm~\ref{alg:correction} summarizes the correction, calibration, and frozen-transfer procedure.

\begin{algorithm}[!t]
\caption{Mesh-free occupancy volume correction with a split-conformal error bound (development, then frozen transfer).}
\label{alg:correction}
\small
\begin{algorithmic}[1]
\Require Development objects $\mathcal{D}$ with query points $\mathbf{x}_{\mathrm{norm}}$, labels, scale $s$, offset $\mathbf{loc}$, and reference mesh volume $V_m$; new batch $\mathcal{D}'$ without meshes; coverage level $1-\alpha=0.95$
\Function{Features}{object}
  \State $\mathbf{x}_{\mathrm{raw}}\gets s\,\mathbf{x}_{\mathrm{norm}}+\mathbf{loc}$ \Comment{unpacked labels}
  \State $B_q\gets$ product of query-coordinate spans; $p\gets$ inside fraction
  \State $V_o\gets pB_q$; $B_{\mathrm{in}}\gets$ bounding-box volume of inside points
  \State \Return $\mathbf{f}=[V_o/B_{\mathrm{in}},\,p,\,\log V_o,\,\log B_{\mathrm{in}},\,\text{span ratio}]$, $V_o$
\EndFunction
\For{each $i\in\mathcal{D}$}
  \State $(\mathbf{f}_i,V_{oi})\gets$ \Call{Features}{$i$}; $e_i\gets100(V_{oi}/V_{mi}-1)$
\EndFor
\State Split $\mathcal{D}$ by mesh-volume decile (seed 2026) into train, calibration, test
\State Standardize $\mathbf{f}$ and fit OLS $\widehat{e}(\mathbf{f})$ on train only
\For{each $i$ in calibration}
  \State $\widehat{e}_i\gets\mathrm{clip}(\widehat{e}(\mathbf{f}_i),-50,100)$; $\widehat V_i\gets V_{oi}/(1+\widehat{e}_i/100)$
  \State $a_i\gets\bigl|100(\widehat V_i/V_{mi}-1)\bigr|$ \Comment{conformity score}
\EndFor
\State $q\gets$ the $\lceil(n_{\mathrm{cal}}+1)(1-\alpha)\rceil$th smallest $a_i$ \Comment{$q=2.661\%$}
\State Freeze and hash scaler, coefficients, clipping rule, and $q$
\For{each $j\in\mathcal{D}'$} \Comment{no refitting, no mesh}
  \State $(\mathbf{f}_j,V_{oj})\gets$ \Call{Features}{$j$}; $\widehat{e}_j\gets\mathrm{clip}(\widehat{e}(\mathbf{f}_j),-50,100)$
  \State $\widehat V_j\gets V_{oj}/(1+\widehat{e}_j/100)$
  \State Report $\bigl[\widehat V_j/(1+q/100),\ \widehat V_j/(1-q/100)\bigr]$
\EndFor
\State Evaluate MdAPE, RMSE, and coverage overall and for $p<0.02$ against $V_m$
\end{algorithmic}
\end{algorithm}

\subsection{Boundary and subgroup diagnostics}

The boundary analysis uses 60 development objects sampled without replacement with seed 790; their recorded IDs are reused in the exact-distance refit. That refit uses query-sampling seed $j$ for zero-based object index $j$. It analyzes the occupancy queries near the raw mesh in normalized coordinates. Points with an approximate distance below 0.02 (computed from the 32 nearest triangle centroids) are candidates, and up to 3,000 per object are retained. Their distances are then computed exactly, as the point-to-triangle distance over every triangle that could lie closer than the approximate value, with the inside/outside sign given by the generalized winding number \cite{ref24}. The labels are fitted with the logistic model
\begin{equation}
P(\mathrm{inside}\mid d)=\left[1+\exp\!\left(\frac{d-t_{50}}{w}\right)\right]^{-1},
\end{equation}
where the midpoint $t_{50}$ estimates the label-transition displacement and $w$ its width. The model is fitted by maximum likelihood from five starting points. A fit is accepted only if the optimizer reports convergence, the scaled projected gradient is below $10^{-4}$, neither parameter lies on a bound, the starts that reach the optimum agree on $t_{50}$ to within $10^{-5}$, each label class contains at least 20 points, and the classes overlap.

In the documented label pipeline \cite{ref20}, \cite{ref34}, meshes are scaled to largest extent 0.9, and depth maps from 100 views, each shifted 1.5 voxels toward the camera, are fused at $256^3$ resolution. E8 tests this pipeline directly.

Second-batch diagnostics compare shape distributions, implied offsets, low-occupancy prevalence, and coverage by shape-complexity decile. A development regression of error on shape complexity predicts the change in mean raw error expected from the change in mean shape complexity; this is an association-based decomposition. The worst-error analysis is post hoc. A zero-intercept model $e=\beta A_ms/V_m$, fitted on development objects and applied unchanged to the second batch, serves as a mechanistic diagnostic; it uses mesh information and is not a deployment correction.

\subsection{Controlled label-pipeline reimplementation}

E8 reimplemented the 3DMSL label step in NumPy from the repository code \cite{ref20}, \cite{ref22}, because the original libraries require compiled graphics components. Each mesh is rendered from the same 100 views ($640\times640$ depth maps); each depth map is shifted toward the camera by the depth offset and eroded with a $3\times3$ minimum filter; and the maps are fused into a truncated signed-distance field at $256^3$ with a truncation of 10 voxels. The stored query points are labeled inside where the interpolated field is negative. For 55 of the 60 E5 objects, labels were generated at depth offsets of 0, 0.75, and 1.5 voxels (1.5 is the dataset default) with nothing else changed (paired within object). The primary endpoint is the change in occupancy-volume error from offset 0 to 1.5, with a 5,000-sample paired object bootstrap interval. Fidelity is measured by agreement with the released labels for sampled query points within 0.01 normalized units of the mesh and by the correlation of recreated and released per-object errors. Midpoint changes use objects whose fits pass the E5 acceptance checks at both offsets. For ten objects, a hybrid labeling replaced interpolated labels with winding-number containment in the extracted mesh for query points whose interpolated field magnitude was below $2/256$, retaining interpolated labels elsewhere, a partial containment check. E8 measures fidelity to the released labels, not identity with the historical build; Supplementary Section~S4 records the pilot exclusion and coordinate-convention choice.

\subsection{Real EM processing}

MitoEM-R analysis uses labels on slices 400--499 at $30\times8\times8$ nm spacing in $z$, $y$, and $x$. Eligible objects contain at least 2,000 voxels, span at least three $z$ planes, do not touch slab or image boundaries, and have a padded region of interest (ROI) of at most 2,000,000 voxels. Of 1,203 instance IDs, sequential filtering removed 586 for slab-boundary contact, 61 for image-boundary contact, none for size or $z$ span, and six for ROI size, leaving 550 (45.7\%). All 550 were meshed by marching cubes \cite{ref23} at level 0.5. Mesh volume and area are compared with voxel-count volume and exposed-voxel-face area. Surface point sampling uses 10,000 points and five repetitions per object. All representations derive from the same annotation, so this is a processing-portability experiment.

In E7, masks are downsampled in $y$ and $x$ only, by block occupancy of at least one half with ties assigned to foreground. Spacings become $30\times16\times16$ and $30\times24\times24$ nm; $z$ is unchanged. The 24 nm in-plane spacing matches the reported 3DMSL mesh-generation spacing; the complete voxel geometry remains $30\times24\times24$ nm \cite{ref20}. Ties can occur only for the $2\times2$ blocks of the 16 nm grid; the $3\times3$ blocks of the 24 nm grid cannot tie. Remeshed volume, surface area, and sphericity, $\pi^{1/3}(6V)^{2/3}/A$, are compared with the native mesh. The experiment measures sensitivity to this regridding and meshing procedure, not the full effect of lower-resolution acquisition. Supplementary analysis S3 repeats it over all block phases and tie rules.

\subsection{Statistical reporting}

We use percentile bootstrap intervals \cite{ref27}. Frozen evaluation uses 1,000 object-level resamples (seed 790); internal validation uses 2,000 resamples (seed 2026 with method-specific offsets). These intervals describe object-level variability and do not capture between-animal uncertainty. The primary uncertainty bound is the saved 2.661\% development-calibrated bound, not recalibrated on the second batch. The 2.17\% margin is the median pairwise repeatability coefficient from E2 and is retained as a descriptive sensitivity threshold, not a single-estimate accuracy bound; 5\%, 6\%, and 10\% are sensitivity thresholds. None is a biological equivalence margin. Subgroup coverage intervals use the 95\% Wilson method as an object-independence approximation.

\section{Results}

\subsection{High overall agreement coexists with systematic representation bias}

All 2,720 development objects were retained. The audit flagged 73 objects for low occupancy fraction, four for a raw point-cloud extent discrepancy above 10\% on at least one axis, and two for incomplete rendered images. Volume ICC was 0.9939, while mean signed occupancy error was $+3.665$\% (bootstrap interval $+3.572$\% to $+3.755$\%) and median signed error was $+3.052$\%. Occupancy volume exceeded mesh volume for 2,719 of 2,720 objects.

The ICC largely reflects the 68-fold range of mesh volumes. Within volume deciles, the median ICC was 0.8570 (range 0.747--0.983). Bland--Altman analysis gave a bias of $+1{,}369$ cubic mesh units ($+4.67$\% of the mean mesh volume), 95\% limits of agreement of $-3{,}568$ to $+6{,}306$, a percentage error of 16.47\% using the mean of both methods (16.85\% using mean mesh volume), and a proportional-bias correlation of 0.82: the absolute discrepancy grows with object size (Supplementary Fig.~S2). Occupancy extents were nearly unbiased ($-0.31$\% to $-0.03$\% by axis), whereas trimmed point-cloud extents were smaller than mesh extents by 1.50\% to 2.31\%, the expected effect of discarding extremal points.

In E2, the median estimated full-sample sampling SD was 0.784\% (interquartile range 0.508\%--1.088\%), giving a median repeatability coefficient of 2.172\%. The median signed error in this 400-object subset was $+2.930$\% at 100,000 queries, about 3.7 times the median sampling SD. This supports a systematic component beyond sampling variability. Under the binomial query model, sampling SD grows at low occupancy, reaching 2.21\% at $p=0.02$.

\subsection{An outward label-boundary offset is observed}

\begin{figure*}[!t]
\centering
\includegraphics[width=\linewidth]{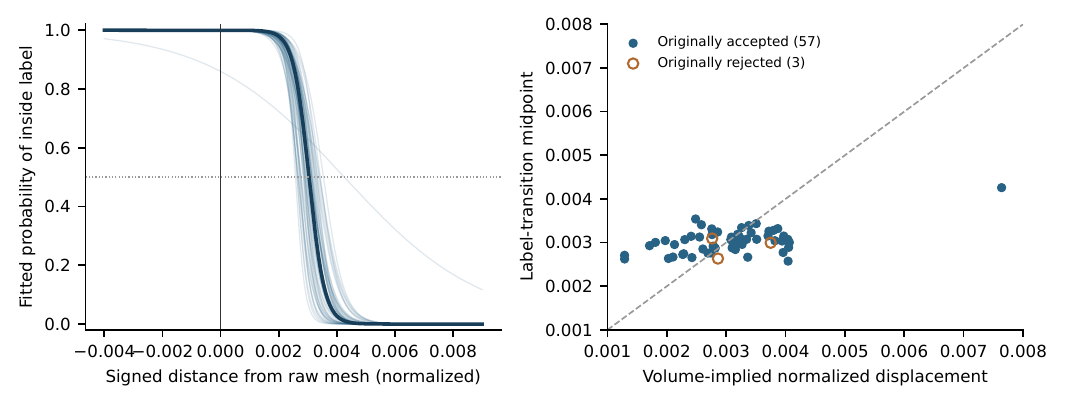}
\caption{Exact-distance boundary fits for the 57 objects accepted in the recorded analysis, replayed on the original queries. Left: object-specific fitted transitions (thin lines) and the transition at median fitted parameters (bold); zero marks the raw mesh. Right: label-transition midpoint versus volume-implied displacement, with identity line ($r=0.58$). Open markers show the three fits that failed the acceptance checks; they are excluded from the correlation.}
\label{fig:mechanism}
\end{figure*}

Of the 60 fits, 57 met every acceptance criterion; the other three stopped with an optimizer termination warning (not on a bound; midpoints 0.00263--0.00309, within the accepted range) and are excluded from primary summaries. For the accepted fits, the median label-transition midpoint was $t_{50}=0.00304$ normalized units outside the mesh (interquartile range 0.00287--0.00315; coefficient of variation 0.09), with a median transition width of 0.00027. The median volume-implied offset was 0.00312, a ratio of 0.98, and the per-object Pearson correlation between the two was 0.58 (Spearman 0.35). One object with a large volume-implied displacement raises the Pearson value (without it, $r=0.31$). The 32-centroid approximation returned the exact distance for every selected point. No point more than ten transition widths inside the boundary was labeled outside, and at most 0.1\% of points more than ten widths outside were labeled inside, supporting a displaced label transition with few far-from-transition inconsistencies (Fig.~\ref{fig:mechanism}).

The labels therefore describe a surface slightly outside the released mesh, consistent with the volume bias arising in dataset generation.

\subsection{The frozen correction transfers with reduced coverage}

Table~\ref{tab:IV} compares internal and frozen second-batch performance. The occupancy-only correction reduced internal MdAPE from 3.035\% to 0.580\%. On the 2,728 new objects, it reduced MdAPE from 3.481\% to 0.664\% and RMSE from 5.743\% to 1.743\%, reductions of 80.9\% and 69.7\%. Its second-batch mean signed bias was $+0.006$\% (95\% bootstrap interval $-0.059$\% to $+0.076$\%), and the MdAPE interval was 0.636\%--0.700\%.

\begin{table}[!t]
\caption{Correction performance. Errors are percentages relative to mesh volume. Coverage uses each method's own calibrated absolute-error bound.}
\label{tab:IV}
\centering
\footnotesize
\setlength{\tabcolsep}{2.5pt}
\renewcommand{\arraystretch}{1.15}
\begin{tabular}{@{}lrrrrr@{}}
\toprule
\textbf{Method} & \textbf{Bias} & \textbf{MdAPE} & \textbf{RMSE} & \textbf{Bound} & \textbf{Cov.} \\
\midrule
\multicolumn{6}{@{}l}{\textit{Internal test, n = 550}} \\
Uncorrected & $+3.608$ & 3.035 & 4.286 & 8.451 & 96.4\% \\
Global correction & $-0.040$ & 1.370 & 2.231 & 4.632 & 96.4\% \\
Occupancy-only & $-0.032$ & 0.580 & 1.149 & 2.661 & 96.2\% \\
\midrule
\multicolumn{6}{@{}l}{\textit{Frozen second batch, n = 2,728}} \\
Uncorrected & $+4.497$ & 3.481 & 5.743 & 8.451 & 90.0\% \\
Global correction & $+0.817$ & 1.503 & 3.542 & 4.632 & 90.0\% \\
Occupancy-only & $+0.006$ & 0.664 & 1.743 & 2.661 & 92.1\% \\
\bottomrule
\end{tabular}
\end{table}

\begin{figure}[!t]
\centering
\includegraphics[width=0.82\linewidth]{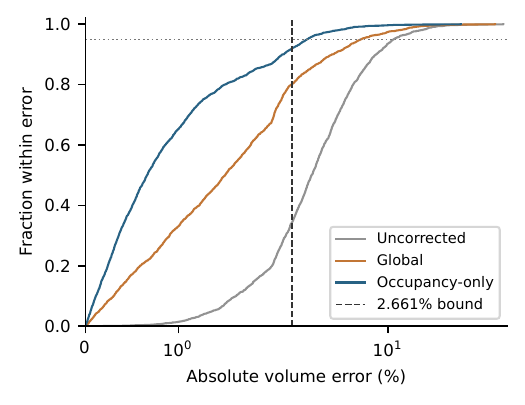}
\caption{Empirical cumulative distributions of absolute percentage error for all 2,728 second-batch objects. Every observation is included. The horizontal axis is linear near zero and logarithmic in the tail. The vertical line is the occupancy-only 2.661\% bound; the horizontal dotted line marks 95\% coverage.}
\label{fig:frozen}
\end{figure}

A near-zero mean residual does not imply uniformly accurate estimates. Coverage of the occupancy-only bound fell to 92.1\% (bootstrap interval 91.1\%--93.0\%). Corrected second-batch pass rates at 2.17\%, 5\%, 6\%, and 10\% were 88.5\%, 98.0\%, 98.8\%, and 99.7\%. Fig.~\ref{fig:frozen} shows the complete absolute-error distributions, including the tails; the vertical line marks the occupancy-only bound. The uncorrected and global methods have identical coverage because they cover the same objects: the global correction divides every estimate by $k=1+c/100$ ($c=3.65$\%, the mean training error), so it covers raw errors from $c-b_gk=-1.15$\% to $c+b_gk=8.451$\%; the upper limit equals the uncorrected bound because both calibration quantiles fall on the same calibration object, and no raw error fell below the lower limit.

\subsection{The label pipeline reproduces the bias, and removing its depth offset removes about half of it}

\normalcolor The reimplementation reproduced the released labels closely. At the default offset, a median of 98.0\% of sampled near-surface query points received the released label, compared with 91.7\% at offset 0 and 95.0\% at 0.75. Across objects, recreated and released volume errors were almost perfectly correlated (Pearson $r=0.999$), although the recreated errors were systematically smaller (median $+3.43$\% versus $+3.90$\%; median per-object difference 0.37 percentage points). In ten objects, replacing near-surface interpolated labels with mesh-containment labels (hybrid labeling) changed volume errors by at most 0.26 percentage points and did not close this gap; complete containment labeling was not tested.

\begin{figure}[!t]
\centering
\includegraphics[width=0.78\linewidth]{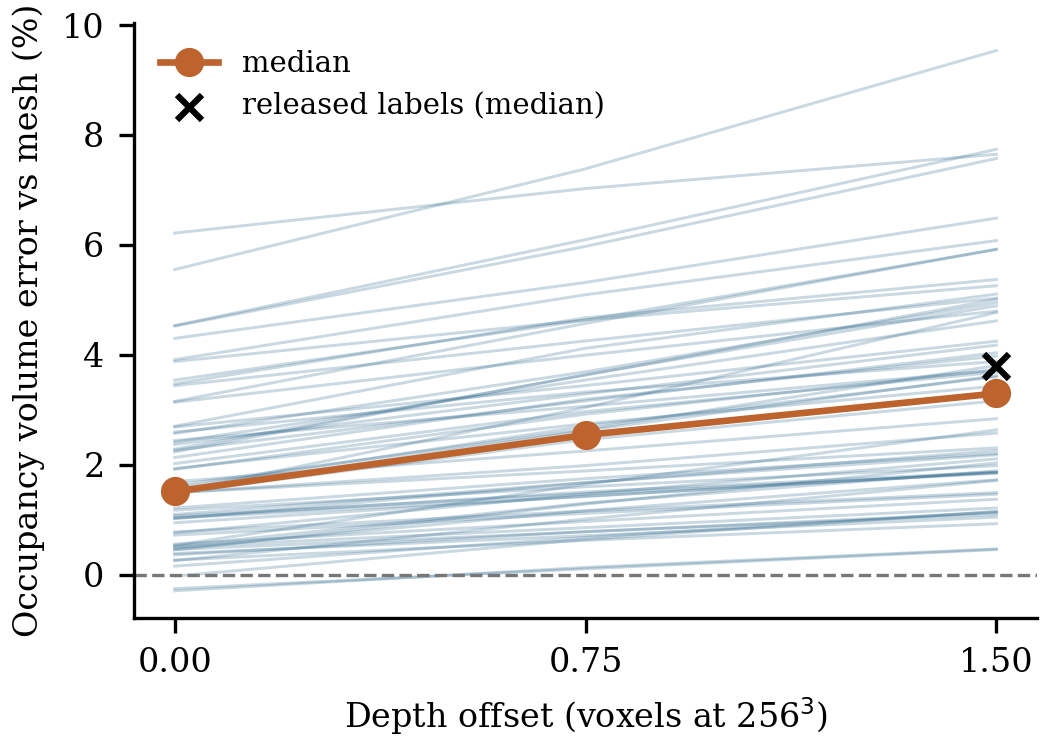}
\caption{Controlled label-pipeline test (E8). Occupancy-volume error relative to the raw mesh at depth offsets of 0, 0.75, and 1.5 voxels; thin lines are individual objects and the cross marks the median error of the released labels. The plot includes the five pilot objects; statistics in the text use the other 55.}
\label{fig:dose}
\end{figure}

Removing the depth offset lowered the occupancy-volume error in every one of the 55 objects, by a mean of 1.57 percentage points (95\% bootstrap interval 1.37--1.77), and the median error rose monotonically from 1.578\% to 2.649\% and 3.435\% across 0, 0.75, and 1.5 voxels (Fig.~\ref{fig:dose}); midpoint fits were accepted for 54 of 55 objects at each offset. The fused-mesh volume error fell by 1.48 points, and in the 53 objects with accepted fits at both offsets the label midpoint moved inward in every object, by a mean of 0.00126. Removing the offset therefore removed about 45\% of the reproduced inflation. The source of the remainder was not isolated: depth-map erosion and truncated-distance fusion are candidate contributors, but neither was varied separately. The alternative coordinate convention and inclusion of the five pilot objects gave the same effect (1.59 and 1.58 points).

\subsection{Shape distribution and low occupancy define the transfer limits}

The second batch contained larger and less spherical objects: median sphericity was 0.793 versus 0.8358, median $A_m/V_m^{2/3}$ was 6.098 versus 5.786, and median occupancy fraction was 0.1138 versus 0.1437. Low-occupancy prevalence increased from 73/2,720 (2.7\%) to 181/2,728 (6.6\%).

Mean raw error increased by 0.833 percentage points; the development regression on shape complexity predicted 0.747 points (about 90\%), a descriptive decomposition. The offset itself replicated: median volume-implied offsets were 0.00293 (development) and 0.00297 (second batch), with coefficients of variation of 0.28 and 0.27, and the slope of error on $A_m/V_m^{2/3}$ was 1.78 and 2.00 percentage points per unit ($R^2=0.67$ and 0.69). The scale-aware predictor $A_ms/V_m$ fitted on development objects (slope 0.310, equivalent to $t_{\mathrm{norm}}=0.00310$) explained the second-batch errors without refitting ($R^2=0.83$; mean residual $-0.06$ points).

\begin{table}[!t]
\caption{Occupancy-only coverage at the saved 2.661\% bound. Wilson intervals assume independent objects and do not quantify between-animal uncertainty; $p$ is the occupancy fraction.}
\label{tab:V}\centering\footnotesize
\setlength{\tabcolsep}{3pt}\renewcommand{\arraystretch}{1.1}
\begin{tabular}{@{}llrrrl@{}}
\toprule
\textbf{Eval.} & \textbf{Group} & \textbf{Covered} & \textbf{Fail.} & \textbf{MdAPE} & \textbf{Coverage [95\% CI]}\\
\midrule
Internal & All & 529/550 & 21 & 0.580\% & 96.2\% [94.2, 97.5]\\
Internal & $p<0.02$ & 7/16 & 9 & 3.011\% & 43.8\% [23.1, 66.8]\\
Internal & $p\ge0.02$ & 522/534 & 12 & 0.551\% & 97.8\% [96.1, 98.7]\\
Second & All & 2,512/2,728 & 216 & 0.664\% & 92.1\% [91.0, 93.0]\\
Second & $p<0.02$ & 89/181 & 92 & 2.766\% & 49.2\% [42.0, 56.4]\\
Second & $p\ge0.02$ & 2,423/2,547 & 124 & 0.625\% & 95.1\% [94.2, 95.9]\\
\bottomrule\end{tabular}\end{table}

Table~\ref{tab:V} identifies much higher failure risk below occupancy 0.02, but the flag does not capture all violations: 124 of 216 second-batch failures (57.4\%) occurred outside it. Unflagged coverage decreased from 97.8\% internally to 95.1\% in the second batch. Reweighting the internal subgroup coverages to second-batch subgroup proportions gives 94.17\% coverage, versus 96.18\% internally and 92.08\% observed. This attributes 2.01 percentage points of the decline to subgroup composition and a further 2.09 points to within-group changes. Using the audit flag as an applicability restriction is a later interpretation, and the observed 95.1\% is not a conditional guarantee. Within the second batch, coverage declined across shape-complexity deciles from 100\% to 64.1\%, and the highest decile remained under-corrected by a mean of $+1.45$\%. The worst 136 errors had median sphericity 0.557 and occupancy 0.017 (cohort medians: 0.793 and 0.114).

\subsection{Real EM processing preserves volume more than surface descriptors}

All 550 eligible MitoEM-R objects were meshed. The median mesh-versus-voxel volume difference was $-0.19$\% (5th--95th percentile $-0.35$\% to $-0.10$\%). The median mesh-versus-voxel-face surface difference was $-16.31$\%, illustrating the dependence of digitized-surface measurements on the area estimator \cite{ref30}. The median trimmed point-cloud extent difference was $-3.70$\%, whereas raw sampled extents differed by only $-0.03$\%, reproducing on real EM the trimming effect seen in 3DMSL.

\begin{table}[!t]
\caption{MitoEM-R in-plane regridding relative to native-grid meshes for the same 550 objects (medians; brackets show 5th--95th percentiles). $\rho$: sphericity rank correlation.}
\label{tab:VI}
\centering
\footnotesize
\setlength{\tabcolsep}{2.5pt}
\renewcommand{\arraystretch}{1.1}
\begin{tabular}{@{}lrrrr@{}}
\toprule
\textbf{Spacing, nm} & \textbf{Volume} & \textbf{Surface area} & \textbf{Sphericity} & $\boldsymbol{\rho}$ \\
\midrule
$30\times16\times16$ & $+0.98$\% & $-5.84$\% & $+6.90$\% & 0.997 \\
 & [$+0.45$, $+1.46$] & [$-7.68$, $-4.72$] & [$+5.68$, $+8.69$] & \\
$30\times24\times24$ & $-0.08$\% & $-10.60$\% & $+11.76$\% & 0.994 \\
 & [$-0.93$, $+0.41$] & [$-12.14$, $-8.67$] & [$+9.24$, $+13.75$] & \\
\bottomrule\end{tabular}\end{table}

All meshes in the original regridding analysis remained watertight. At the original 24 nm in-plane grid phase, volume rank correlation was 1.000 to reported precision, while median sphericity increased by 11.76\% (Table~\ref{tab:VI}). The near-zero volume change at 24 nm holds across grid alignments, but the 16 nm result does not: across all block phases and tie rules (Supplementary Section~S3), median volume changes ranged from $-0.22$\% to $+0.01$\% at 24 nm but from $-5.47$\% to $+6.94$\% at 16 nm. Euler characteristics changed in nine of 550 objects at each spacing, so watertightness does not guarantee topology preservation. These shifts are not correction factors for other tissues or instruments.

\section{Discussion}

\subsection{Measurement agreement is a prerequisite for biological interpretation}

The experiments separate three properties that are easily conflated. An estimator can be numerically precise while targeting a boundary that differs from the reference mesh (sampling SD about 0.8\%, systematic offset nearly four times larger); a correction can remove average bias while leaving substantial errors in a morphological subgroup; and object rankings can remain stable while a descriptor's absolute scale changes. The high ICC, the near-zero corrected mean bias, and the high regridding rank correlations are each informative, but none alone establishes interchangeability.

For users of 3DMSL and similar resources, three recommendations follow. First, occupancy labels describe a surface about 0.003 normalized units outside the released mesh, so derived volumes and models trained on them inherit a bias whose relative size scales with $A_ms/V_m$. Second, in our reimplementation, zero depth offset removed only about half of this bias; labeling query points directly against the released mesh avoids the fusion step. Third, a mesh-free correction reduces typical error about fivefold, but its uncertainty bound should not be relied on for objects with occupancy fraction below 0.02, and it does not guarantee coverage for other objects either.

\subsection{Correction and uncertainty transfer are separate results}

Observed coverage was near 95\% in the unflagged group and about 49\% in the flagged group, and it declined within the unflagged group between evaluations, illustrating the distinction between marginal and conditional coverage \cite{ref29}. Conditional residual scales, Mondrian calibration by occupancy group, or covariate-shift weighting \cite{ref28} are natural next steps, but they require new development, and the second batch could not serve as an untouched test for methods selected using its outcomes.

One potential application is Homeostatic Invariance, which we propose as a framework in which structural variability may coexist with maintained function within a bounded biological regime. The findings address a measurement prerequisite for testing that proposition, not the proposition itself. Such a test would need aligned functional observations, longitudinal or perturbational designs, and replication across biological units; the present datasets contain no linked respiration, ATP, or membrane-potential measurements.

\section{Limitations}

The two 3DMSL batches share construction and provenance; disjoint IDs do not establish independent acquisitions or animals. Object-level bootstrap intervals may be optimistic under spatial or biological dependence. The MitoEM-R analysis covers one rat-cortex volume and excludes boundary-truncated, small, shallow, and very large objects. The internal split followed exploratory analysis, and E8 and the supplementary analyses were added after the primary results had been inspected (Supplementary Material, Section~S4). E8 is a reimplementation whose correspondence to the historical code has not been confirmed with the dataset authors; it under-predicts the released errors by a median of 0.37 percentage points, labels points from the interpolated field, and does not separate erosion from fusion. No new-source evaluation of the correction was performed. The regridding experiment leaves $z$ unchanged and does not model the point-spread function, segmentation uncertainty, or another instrument's acquisition. Candidate tolerances are not biological equivalence margins.

\section{Conclusion}

This study establishes four empirical findings for quantitative mitochondrial imaging: high whole-sample agreement can conceal a systematic representation bias; a controlled processing change can reduce that bias without eliminating it; correction accuracy can transfer while calibrated coverage deteriorates; and descriptor rankings can remain stable while absolute measurements shift. The resulting validation requirements are concrete: report signed object-level errors alongside agreement coefficients, test correction accuracy and uncertainty coverage separately on held-out objects, and assess grid alignment and tie rules before interpreting descriptor changes. The reported effect sizes apply to the tested resources and procedures and are not universal correction factors; biological function and cross-source correction transfer remain for future validation.

\section*{Data, Code, and Ethics}

The source datasets, 3DMSL \cite{ref12} and MitoEM \cite{ref15}, are publicly available. The reproducibility code and supporting materials, including executed notebooks, processed per-object results, split identifiers, and the frozen model, are publicly available at \url{https://github.com/farouze/mitocondria}. This study analyzed only publicly available, previously published datasets. No new animal experiments or human-participant research were conducted.

\bibliographystyle{IEEEtran}
\bibliography{references}

@article{ref1,
  author = {A. M. van der Bliek and Q. Shen and S. Kawajiri},
  title = {{Mechanisms of mitochondrial fission and fusion}},
  journal = {Cold Spring Harbor Perspectives in Biology},
  volume = {5},
  year = {2013},
  number = {6},
  pages = {a011072},
  url = {https://doi.org/10.1101/cshperspect.a011072},
  doi = {10.1101/cshperspect.a011072}
}

@article{ref2,
  author = {G. Benard and others},
  title = {{Mitochondrial bioenergetics and structural network organization}},
  journal = {Journal of Cell Science},
  volume = {120},
  year = {2007},
  pages = {838--848},
  url = {https://doi.org/10.1242/jcs.03381},
  doi = {10.1242/jcs.03381}
}

@article{ref3,
  author = {J. Faitg and T. Davey and D. M. Turnbull and K. White and A. E. Vincent},
  title = {{Mitochondrial morphology and function: two for the price of one!}},
  journal = {Journal of Microscopy},
  volume = {278},
  year = {2020},
  pages = {89--106},
  url = {https://eprints.ncl.ac.uk/266059},
  doi = {10.1111/jmi.12891}
}

@article{ref4,
  author = {W.-W. Tseng and others},
  title = {{Metabolic regulation of mitochondrial morphologies in pancreatic beta cells: coupling of bioenergetics and mitochondrial dynamics}},
  journal = {Communications Biology},
  volume = {7},
  year = {2024},
  pages = {1267},
  url = {https://doi.org/10.1038/s42003-024-06955-3},
  doi = {10.1038/s42003-024-06955-3}
}

@article{ref5,
  author = {Y. Ding and others},
  title = {{Single Mitochondrion Morphology-Function Relationship Analysis Using Fluorescent Probes and Artificial Intelligence}},
  journal = {Advanced Science},
  volume = {12},
  year = {2025},
  number = {43},
  pages = {e09140},
  url = {https://doi.org/10.1002/advs.202509140},
  doi = {10.1002/advs.202509140}
}

@article{ref6,
  author = {S. Charrasse and others},
  title = {{Combining mitochondrial morphology and cellular metabolism measurements improves colorectal cancer cell classification}},
  journal = {Advances in Cancer Biology - Metastasis},
  volume = {16},
  year = {2026},
  pages = {100175},
  url = {https://doi.org/10.1016/j.adcanc.2026.100175},
  doi = {10.1016/j.adcanc.2026.100175}
}

@article{ref7,
  author = {O. Tovar-Ferrero and J. Rubio and A. Zorzano and G. Martinez-Corrales and M. Liesa},
  title = {{Measuring mitochondrial membrane potential}},
  journal = {The EMBO Journal},
  volume = {44},
  year = {2025},
  pages = {7334--7345},
  url = {https://doi.org/10.1038/s44318-025-00632-9},
  doi = {10.1038/s44318-025-00632-9}
}

@article{ref9,
  author = {S. Zaghbani and L. Faber and A. J. Garcia-Saez},
  title = {{Convolutional Neural Network approach to classify mitochondrial morphologies}},
  journal = {Computational Biology and Chemistry},
  volume = {118},
  year = {2025},
  pages = {108477},
  url = {https://doi.org/10.1016/j.compbiolchem.2025.108477},
  doi = {10.1016/j.compbiolchem.2025.108477}
}

@article{ref10,
  author = {A. Kaianathbhatta and others},
  title = {{MitoTex (Mitochondria Texture Analysis User Interface): Open-Source Framework for Textural Characterization and Classification of Mitochondrial Structures}},
  journal = {International Journal of Molecular Sciences},
  volume = {27},
  year = {2026},
  number = {3},
  pages = {1191},
  url = {https://doi.org/10.3390/ijms27031191},
  doi = {10.3390/ijms27031191}
}

@article{ref11,
  author = {A. R. Punnakkal and S. S. Jadhav and A. V. Celeste and A. Horsch and K. Agarwal and D. K. Prasad},
  title = {{3D Mitochondria Shape Library for Optical Microscopy (3DMSL): A multimodal dataset for deep learning based mitochondrial analysis}},
  journal = {Data in Brief},
  volume = {65},
  year = {2026},
  pages = {112507},
  url = {https://doi.org/10.1016/j.dib.2026.112507},
  doi = {10.1016/j.dib.2026.112507}
}

@misc{ref12,
  author = {A. R. Punnakkal and others},
  title = {{3DMSL - 3D Mitochondria Shape Library for Optical Microscopy}},
  note = {DataverseNO, dataset, version 2, 2024},
  howpublished = {\url{https://doi.org/10.18710/JX6JXF}},
  doi = {10.18710/JX6JXF}
}

@misc{ref13,
  author = {G. Parlakg{\"u}l and G. S. Hotamisligil},
  title = {{High resolution 3D imaging of liver subcellular architecture and its link to metabolic function}},
  note = {EMPIAR-10791, dataset, released 2022, accessed Sep. 25, 2026},
  howpublished = {\url{https://doi.org/10.6019/EMPIAR-10791}},
  doi = {10.6019/EMPIAR-10791}
}

@article{ref14,
  author = {G. Parlakg{\"u}l and others},
  title = {{Regulation of liver subcellular architecture controls metabolic homeostasis}},
  journal = {Nature},
  volume = {603},
  year = {2022},
  pages = {736--742},
  url = {https://doi.org/10.1038/s41586-022-04488-5},
  doi = {10.1038/s41586-022-04488-5}
}

@inproceedings{ref15,
  author = {D. Wei and others},
  title = {{MitoEM Dataset: Large-scale 3D Mitochondria Instance Segmentation from EM Images}},
  booktitle = {Proc. Med. Image Comput. Comput. Assist. Interv. (MICCAI)},
  series = {Lecture Notes in Computer Science},
  volume = {12265},
  pages = {66--76},
  year = {2020},
  url = {https://doi.org/10.1007/978-3-030-59722-1_7},
  doi = {10.1007/978-3-030-59722-1_7}
}

@article{ref16,
  author = {D. Franco-Barranco and others},
  title = {{Current Progress and Challenges in Large-Scale 3D Mitochondria Instance Segmentation}},
  journal = {IEEE Transactions on Medical Imaging},
  volume = {42},
  year = {2023},
  number = {12},
  pages = {3956--3971},
  url = {https://doi.org/10.1109/TMI.2023.3320497},
  doi = {10.1109/TMI.2023.3320497}
}

@article{ref17,
  author = {J. M. Bland and D. G. Altman},
  title = {{Statistical methods for assessing agreement between two methods of clinical measurement}},
  journal = {The Lancet},
  volume = {327},
  year = {1986},
  number = {8476},
  pages = {307--310},
  url = {https://doi.org/10.1016/S0140-6736(86)90837-8},
  doi = {10.1016/S0140-6736(86)90837-8}
}

@article{ref18,
  author = {T. K. Koo and M. Y. Li},
  title = {{A Guideline of Selecting and Reporting Intraclass Correlation Coefficients for Reliability Research}},
  journal = {Journal of Chiropractic Medicine},
  volume = {15},
  year = {2016},
  number = {2},
  pages = {155--163},
  url = {https://doi.org/10.1016/j.jcm.2016.02.012},
  doi = {10.1016/j.jcm.2016.02.012}
}

@misc{ref19,
  author = {A. N. Angelopoulos and S. Bates},
  title = {{A Gentle Introduction to Conformal Prediction and Distribution-Free Uncertainty Quantification}},
  note = {arXiv:2107.07511, 2021},
  howpublished = {\url{https://arxiv.org/abs/2107.07511}}
}

@misc{ref20,
  author = {BioAILab},
  title = {{3DMSL}},
  note = {Official dataset code repository, accessed Sep. 24, 2026},
  howpublished = {\url{https://github.com/bioailab/3DMSL}}
}

@inproceedings{ref21,
  author = {L. Mescheder and M. Oechsle and M. Niemeyer and S. Nowozin and A. Geiger},
  title = {{Occupancy Networks: Learning 3D Reconstruction in Function Space}},
  booktitle = {Proc. IEEE/CVF Conf. Comput. Vis. Pattern Recognit. (CVPR)},
  pages = {4460--4470},
  year = {2019},
  url = {https://openaccess.thecvf.com/content_CVPR_2019/html/Mescheder_Occupancy_Networks_Learning_3D_Reconstruction_in_Function_Space_CVPR_2019_paper.html},
  doi = {10.1109/CVPR.2019.00459}
}

@article{ref22,
  author = {D. Stutz and A. Geiger},
  title = {{Learning 3D Shape Completion Under Weak Supervision}},
  journal = {International Journal of Computer Vision},
  volume = {128},
  pages = {1162--1181},
  year = {2020},
  url = {https://doi.org/10.1007/s11263-018-1126-y},
  doi = {10.1007/s11263-018-1126-y}
}

@article{ref23,
  author = {W. E. Lorensen and H. E. Cline},
  title = {{Marching cubes: A high resolution 3D surface construction algorithm}},
  journal = {ACM SIGGRAPH Computer Graphics},
  volume = {21},
  number = {4},
  pages = {163--169},
  year = {1987},
  url = {https://www.cs.toronto.edu/~jacobson/seminar/lorenson-and-cline-1987.pdf},
  doi = {10.1145/37402.37422}
}

@article{ref24,
  author = {A. Jacobson and L. Kavan and O. Sorkine-Hornung},
  title = {{Robust inside-outside segmentation using generalized winding numbers}},
  journal = {ACM Transactions on Graphics},
  volume = {32},
  number = {4},
  pages = {33:1--33:12},
  year = {2013},
  url = {https://igl.ethz.ch/projects/winding-number/},
  doi = {10.1145/2461912.2461916}
}

@article{ref25,
  author = {K. O. McGraw and S. P. Wong},
  title = {{Forming inferences about some intraclass correlation coefficients}},
  journal = {Psychological Methods},
  volume = {1},
  number = {1},
  pages = {30--46},
  year = {1996},
  url = {https://doi.org/10.1037/1082-989X.1.1.30},
  doi = {10.1037/1082-989X.1.1.30}
}

@article{ref26,
  author = {L. A. H. Critchley and J. A. J. H. Critchley},
  title = {{A meta-analysis of studies using bias and precision statistics to compare cardiac output measurement techniques}},
  journal = {Journal of Clinical Monitoring and Computing},
  volume = {15},
  number = {2},
  pages = {85--91},
  year = {1999},
  url = {https://doi.org/10.1023/A:1009982611386},
  doi = {10.1023/A:1009982611386}
}

@book{ref27,
  author = {B. Efron and R. J. Tibshirani},
  title = {{An Introduction to the Bootstrap}},
  publisher = {Chapman \& Hall},
  address = {New York, NY, USA},
  year = {1993},
  url = {https://www.routledge.com/An-Introduction-to-the-Bootstrap/Efron-Tibshirani/p/book/9780412042317}
}

@inproceedings{ref28,
  author = {R. J. Tibshirani and R. Foygel Barber and E. J. Cand{\`e}s and A. Ramdas},
  title = {{Conformal Prediction Under Covariate Shift}},
  booktitle = {Advances in Neural Information Processing Systems},
  volume = {32},
  year = {2019},
  url = {https://papers.neurips.cc/paper_files/paper/2019/hash/8fb21ee7a2207526da55a679f0332de2-Abstract.html}
}

@article{ref29,
  author = {V. Vovk},
  title = {{Conditional validity of inductive conformal predictors}},
  journal = {Machine Learning},
  volume = {92},
  number = {2--3},
  pages = {349--376},
  year = {2013},
  url = {https://doi.org/10.1007/s10994-013-5355-6},
  doi = {10.1007/s10994-013-5355-6}
}

@article{ref30,
  author = {J. Lindblad},
  title = {{Surface area estimation of digitized 3D objects using weighted local configurations}},
  journal = {Image and Vision Computing},
  volume = {23},
  number = {2},
  pages = {111--122},
  year = {2005},
  url = {https://doi.org/10.1016/j.imavis.2004.06.012},
  doi = {10.1016/j.imavis.2004.06.012}
}

@article{ref31,
  author = {A. J. Valente and L. A. Maddalena and E. L. Robb and F. Moradi and J. A. Stuart},
  title = {{A simple ImageJ macro tool for analyzing mitochondrial network morphology in mammalian cell culture}},
  journal = {Acta Histochemica},
  volume = {119},
  number = {3},
  pages = {315--326},
  year = {2017},
  url = {https://doi.org/10.1016/j.acthis.2017.03.001},
  doi = {10.1016/j.acthis.2017.03.001}
}

@incollection{ref32,
  author = {M. P. Viana and S. Lim and S. M. Rafelski},
  title = {{Quantifying mitochondrial content in living cells}},
  booktitle = {Methods in Cell Biology},
  publisher = {Academic Press},
  volume = {125},
  pages = {77--93},
  year = {2015},
  url = {https://doi.org/10.1016/bs.mcb.2014.10.003},
  doi = {10.1016/bs.mcb.2014.10.003}
}

@article{ref33,
  author = {A. E. Y. T. Lefebvre and others},
  title = {{Nellie: automated organelle segmentation, tracking and hierarchical feature extraction in 2D/3D live-cell microscopy}},
  journal = {Nature Methods},
  volume = {22},
  pages = {751--763},
  year = {2025},
  url = {https://doi.org/10.1038/s41592-025-02612-7},
  doi = {10.1038/s41592-025-02612-7}
}

@misc{ref34,
  author = {D. Stutz},
  title = {{mesh-fusion: Depth rendering and volumetric fusion implementation}},
  note = {Source file 2\_fusion.py, accessed Sep. 25, 2026},
  howpublished = {\url{https://github.com/davidstutz/mesh-fusion/blob/master/2_fusion.py}}
}
\end{document}

\title{Supplementary Material for ``Representation Bias, Correction Transfer, and Resolution Sensitivity in Three-Dimensional Mitochondrial Morphometry''}
\author{\mbox{Farouk Ganiyu Adewumi} and \mbox{Timothy Oladunni}}
\maketitle

This document contains the supplementary analyses (S1--S3) and the analysis history (S4) referenced in the main article.

\section{Supplementary Imaging Analysis}

We tested whether 2D half-peak threshold morphometry of the released 3DMSL simulated microscopy images recovers object geometry (300 development objects, seed 790; Conf1 confocal, 70 nm pixels; Conf2 confocal, 48.8 nm; Epi1 widefield, 109 nm; six orientations; 5,398 views, none empty or edge-touching). The results were negative. Geometry explained only 13\%--28\% of the orientation-to-orientation variation in image area; cross-configuration area ICC(A,1) was 0.509 (Conf1 vs.\ Conf2) and 0.051 and 0.058 (Epi1 pairs); and image-to-geometry area ratios were about 0.77 (confocal) and 2.82 (Epi1), leaving simulator-configuration provenance unresolved. The images were noisy (median signal-to-noise ratio 6--18), objects spanned only a few pixels near the resolution limit (124--245 nm), and hotspots and background noise defeated half-peak and Otsu thresholds. This does not establish a limitation of all segmentation methods; the images were not used in the main analyses.

\section{Supplementary Correction-Model Comparison}

After the original outcomes had been inspected, we compared the retained five-feature ordinary least-squares (OLS) correction with occupancy-fraction-only OLS and with ridge regression on the same features, each fitted on the saved training split with bounds from the calibration split. Ridge used 36 log-spaced penalties from $10^{-3}$ to $10^4$, selected by mean squared error over five unshuffled folds with within-fold standardization ($\alpha=0.251$). Refitting the retained model reproduced the stored residuals (within $4.5\times10^{-14}$ points). Relative to retained OLS, second-batch MdAPE was higher by 0.839 percentage points for the global correction (95\% paired object-bootstrap interval 0.782--0.889), 0.263 for fraction-only OLS (0.221--0.305), and 0.001 for ridge ($-0.003$ to 0.006). Fraction-only OLS had second-batch MdAPE 0.927\%, RMSE 2.718\%, and coverage 89.37\%; ridge had 0.665\%, 1.747\%, and 91.97\%, versus 0.664\%, 1.743\%, and 92.08\% for retained OLS. This is not a formal equivalence test.

\section{Supplementary Grid-Phase Sensitivity}\label{app:phase}

\begin{figure*}[!t]
\centering
\begin{minipage}[b]{0.42\textwidth}\centering
\includegraphics[width=\linewidth]{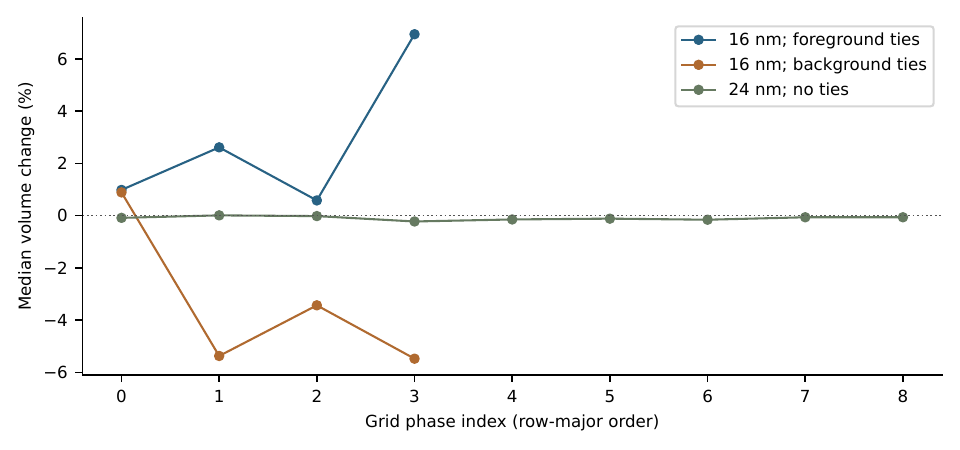}
\caption{Exploratory grid-phase sensitivity for all 550 eligible MitoEM-R objects. Each point is the median volume change from the native-grid mesh. Phase indices enumerate $(y,x)$ offsets in row-major order from $(0,0)$: four phases for factor two, nine for factor three. Leading zero padding changes the block origin without cropping. Lines connect discrete configurations for readability.}
\label{fig:phase}
\end{minipage}\hfill
\begin{minipage}[b]{0.555\textwidth}\centering
\includegraphics[width=\linewidth]{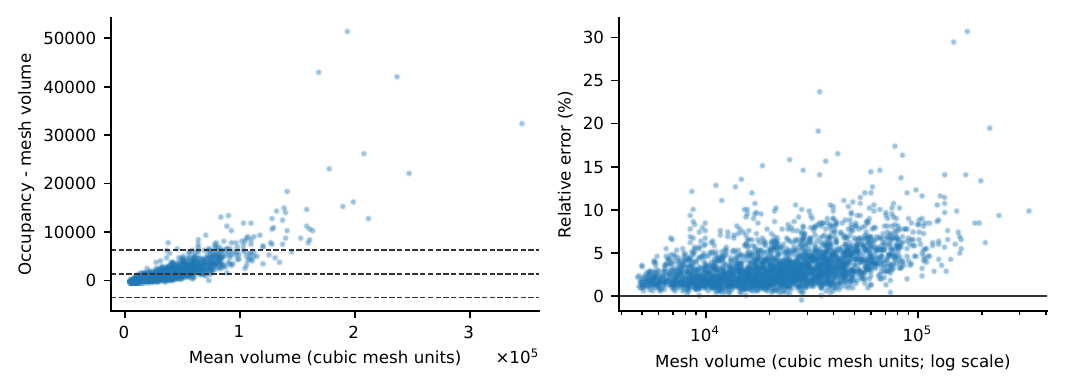}
\caption{Development-batch agreement diagnostics for all 2,720 objects. Left: absolute differences versus pair means, with pooled bias and limits of agreement. Right: relative errors versus mesh volume. The size-dependent pattern limits interpretation of pooled absolute limits as universal error bounds.}
\label{fig:agreement}
\end{minipage}
\end{figure*}

E7 was repeated for all four in-plane block phases at factor two (ties to foreground or background) and all nine at factor three: 17 configurations and 9,350 remeshed pairs, all nonempty (Fig.~\ref{fig:phase}). The original phase begins at the lower bound of each cropped object ROI. Foreground component counts use 6- and 26-neighbor connectivity, and Euler characteristics use 26-neighbor connectivity. At 16 nm, median volume changes ranged from $+0.58$\% to $+6.94$\% across phases with foreground ties and from $-5.47$\% to $+0.89$\% with background ties. At 24 nm they ranged from $-0.22$\% to $+0.01$\%. At the original phase, 26-neighbor component counts changed in one of 550 objects at each spacing.

\section{Analysis History and Deviations}\label{app:history}

\emph{Low-occupancy flag.} The flag ($p<0.02$) was part of the initial-audit quality-control policy; the QC output recording 73 flagged development objects predates the correction. Its use as an applicability restriction is a post hoc interpretation.

\emph{Development and frozen evaluation.} Whole-batch exploration preceded the internal split. The frozen evaluation used the archived model without refitting; all five protocol-listed input hashes match, and the archived model reproduces the stored residuals to within $7\times10^{-14}$ percentage points.

\emph{Boundary fits (E5).} In a later replay all 60 fits converged, giving median $t_{50}=0.00304$ and $r=0.57$. The influential object (25660; without it Spearman $\rho$ falls from 0.351 to 0.315) was not excluded. The geometric expectation of half the depth shift ($0.5\times1.5/(256\times0.9)=0.00326$) is 1.07 times the measured midpoint, but E8 shows the depth offset alone moves the midpoint by only 0.00126, so this ratio is not quantitative agreement.

\emph{Label-pipeline reimplementation (E8).} The first five E5 objects (24558, 24611, 24853, 24898, 24925) were run as an engineering pilot and, inadvertently, again in the full run, with identical results; they were excluded after the full results had been inspected (all 60: 1.58 points, 1.39--1.77). Because the repository code places marching-cubes vertices half a voxel from voxel centers, labels were computed with and without this shift. The selection rule (higher agreement with released labels) was written into the analysis script, whose SHA-256 prefix 30b237e2 was recorded at 16:12 UTC on 26 September 2026, before the runs; the choice was made from full-run results (agreement 98.0\% centered vs.\ 90.2\% shifted; offset effect 1.57 vs.\ 1.59 points). Midpoint acceptance flags were saved without failure reasons, and acceptance for released-label fits was not saved. In the ten-object hybrid containment check (offsets 0 and 1.5), the offset effect was 1.57 points with hybrid labels and 1.68 with interpolated labels.

\emph{Model comparison (S2).} The first ridge run standardized before cross-validation and selected by mean fold $R^2$; the rerun standardizes within folds and selects by mean squared error, so the penalty change ($0.398$ to $0.251$) is not attributable to scaling alone. Displayed metrics changed by at most 0.03 percentage points.

\emph{Other checks.} Using the nominal box width $1.1s$ raises occupancy volume by a median 0.107\% in each batch. The grid-phase checks (S3) and the scale-aware diagnostic were added after the audit and were not prospectively registered.